\documentclass[letterpaper,twoside,12pt, table]{article}
\usepackage{amssymb,amsfonts}
\usepackage{amsmath,amsfonts,bm}
\usepackage{tikz}
\usepackage{graphics}
\usepackage{listings}
\usepackage{subcaption}
\usepackage{stackengine}

\usepackage{latexsym}
\usepackage{graphicx}
\usepackage{mathrsfs}
\usepackage{geometry}
\usepackage{fancyhdr}
\usepackage{booktabs}
\usepackage{multirow}
\usepackage{array}
\usepackage{natbib}
\usepackage[font={footnotesize,bf,sf}]{caption}
\usepackage{color}
\definecolor{gris}{rgb}{0.44,0.44,0.44}
\graphicspath{{Figures/}}
\PassOptionsToPackage{hyphens}{url}
\usepackage{hyperref} 
\hypersetup{colorlinks,%
	citecolor={blue}, 
	urlcolor={blue},
	linkcolor={blue},
	breaklinks={true},
}

\makeatletter
\renewcommand{\section}{\@startsection {section}{1}{\z@}%
             {-2ex \@plus -1ex \@minus -.2ex}%
             {1ex \@plus.2ex}%
             {\normalfont\Large\sffamily\bfseries}}
\renewcommand{\subsection}{\@startsection{subsection}{2}{\z@}%
             {-1.25ex\@plus -1ex \@minus -.2ex}%
             {.75ex \@plus .2ex}%
             {\normalfont\large\sffamily\bfseries}}
\renewcommand{\subsubsection}{\@startsection{subsubsection}{3}%
             {\z@}%
             {-1.25ex\@plus -1ex \@minus -.2ex}%
             {.75ex \@plus .2ex}%
             {\normalfont\normalsize\sffamily\bfseries}}
\renewcommand\paragraph{\@startsection{paragraph}{4}{\z@}%
                                    {-1.25ex \@plus 1ex \@minus -.2ex}%
                                    {-.5em \@plus -.1em}%
                                    {\normalfont\normalsize\sffamily\bfseries}}
\def\@listI{\leftmargin\leftmargini    
            \parsep .25ex \@plus .1ex  
            \topsep .25ex \@plus .1ex  
            \itemsep \parsep}
\let\@listi\@listI
\@listi

\DeclareRobustCommand{\qed}{%
  \ifmmode \mathqed
  \else
    \leavevmode\unskip\penalty9999 \hbox{}\nobreak\hfill
    \quad\hbox{\qedsymbol}%
  \fi
}
\let\QED@stack\@empty
\let\qed@elt\relax
\newcommand{\pushQED}[1]{%
  \toks@{\qed@elt{#1}}\@temptokena\expandafter{\QED@stack}%
  \xdef\QED@stack{\the\toks@\the\@temptokena}%
}
\newcommand{\popQED}{%
  \begingroup\let\qed@elt\popQED@elt \QED@stack\relax\relax\endgroup
}
\def\popQED@elt#1#2\relax{#1\gdef\QED@stack{#2}}
\newcommand{\qedhere}{%
  \begingroup \let\mathqed\math@qedhere
    \let\qed@elt\setQED@elt \QED@stack\relax\relax \endgroup
}
\newcommand{\openbox}{\leavevmode
  \hbox to.77778em{%
  \hfil\vrule
  \vbox to.675em{\hrule width.6em\vfil\hrule}%
  \vrule\hfil}}
\DeclareRobustCommand{\textsquare}{%
  \begingroup \usefont{U}{msa}{m}{n}\thr@@\endgroup
}
\providecommand{\qedsymbol}{\openbox}

\providecommand{\proofname}{\sffamily Proof}

\makeatother

\allowdisplaybreaks

\newlength\llength
\def\eqref#1{equation~\ref{#1}}

\def\1{\bm{1}}

\def\eps{{\epsilon}}

\DeclareMathAlphabet{\mathsfit}{\encodingdefault}{\sfdefault}{m}{sl}
\SetMathAlphabet{\mathsfit}{bold}{\encodingdefault}{\sfdefault}{bx}{n}

\newcommand{\E}{\mathbb{E}}

\DeclareMathOperator*{\argmin}{arg\,min}

\usepackage{tikz,tabu}

\usepackage{cmbright}
\definecolor{green}{rgb}{0.0,0.50,0.0}

\usepackage{amsfonts}
\usepackage{amsmath}
\usepackage{amssymb}
\usepackage{natbib}
\usepackage{dsfont}
\usepackage{graphicx}

\newcommand{\Dc}{\mathcal{D}}
\newcommand{\Sc}{\mathcal{S}}
\newcommand{\Tc}{\mathcal{T}}
\newcommand{\Xc}{\mathcal{X}}

\newcommand{\Hc}{\mathcal{F}}
\newcommand{\Eb}{\mathbb{E}}
\newcommand{\Rf}{\mathfrak{R}}
\newcommand{\one}{\mathds{1}}
\newcommand{\xv}{\mathbf{x}}

\def\leftside{\copyright Huawei Technologies}
\def\rightside{Montreal Research Centre}

\def\shorttitle{Mathematical Challenges in Deep Learning }

\title{\hrule Mathematical Challenges in Deep Learning}

\author{Vahid Partovi~Nia{\footnotesize[1]}, Guojun Zhang{\footnotesize[1]}, Ivan Kobyzev{\footnotesize[1]}, \\ Michael R. Metel{\footnotesize[1]},  Xinlin Li{\footnotesize[1]}, Ke Sun {\footnotesize[3]}, Sobhan Hemati {\footnotesize[1]},\\ Masoud Asgharian{\footnotesize[2]}, Linglong Kong{\footnotesize[3]}, Wulong Liu{\footnotesize[1]}, Boxing Chen{\footnotesize[1]}\footnote{[1] Noah's Ark Lab, [2] McGill University, [3] University of Alberta. This document reflects a subjective viewpoint of the Noah's Ark Montreal Research Centre about some important mathematical challenges in deep learning. The corresponding author is \texttt{boxing.chen@huawei.com}}
}
\date{\parbox{\linewidth}{\centering%
  Open Letter \endgraf\bigskip
 \endgraf \medskip
   \endgraf
 }}

\begin{document}
{\maketitle}

\hrule




\begin{center}
{\textbf {Summary}} 
\end{center}

\noindent Deep models are dominating the artificial intelligence (AI) industry since the ImageNet challenge in 2012. The size of deep models is increasing ever since, which brings new challenges to this field with applications in cell phones, personal computers, autonomous cars, and wireless base stations. Here we list a set of problems, ranging from training, inference, generalization bound, and optimization with some formalism to communicate these challenges with mathematicians, statisticians, and theoretical computer scientists. This is a subjective view of the research questions in deep learning that benefits the tech industry in long run.

\noindent\textbf{Keywords: } Learnable class; Low bit computation; Floating-point arithmetic; Degrees of freedom; Regularization; VC dimension; computational complexity; stochastic gradient descent.
\newpage
\tableofcontents
\newpage

\section{Introduction}
Deep learning-based technology is finding its way to consumer products faster than expected. Conversational agents such as ChatGPT, deep learning-based perception modules in autonomous driving, automatic speech recognition in voice assistants implemented in our cell phones, context-aware translation engines on the web, are all concrete examples. The deep learning community has been obsessed with increasing the accuracy of the model to beat human precision. This started with the ImageNet classification challenge, and growing towards other applications ever since. This obsession with accuracy has led to large models with too many parameters that consequently face two major challenges: i) models are too large that no one can train them anymore, except big enterprises ii) even if the trained model is available, their deployment still relies on big enterprises, due to their large deployment resource requirement. 

This trend will lead to the monopoly of artificial intelligence (AI) innovation to a handful of big enterprises, marginalizing small enterprises, universities, and the public from contributing to this growing field. This trend not only slows down AI innovation  but it may affect AI to serve humanity in long run. We believe a fundamental rethinking of the current research directions is required to address the aforementioned two major issues. There has been efforts to gather important questions of the field such as \cite{dhar2021survey}. New directions has been proposed by fundamental re-thinking about deep models, see for instance \cite{bengio2017consciousness, bengio2021flow, xia2021causal}. We take, however, another perspective in this document and aim to encourage researchers to attack questions that revolve around solving i) and ii) in particular. 

\section{Background}
As models get larger, more memory and computational resources are required to learn (training step) and deploy (inference step) them in practice. We specifically target deep learning models that are emerging fast and transforming the tech industry. We begin by setting the required mathematical notation in Table~\ref{tab:notation}.
\begin{table}[h!]
    \centering
    \begin{tabular}{c l}
    Notation & Description\\ 
    \hline
        $(\mathbf x, y)$ & Observed data, including the input feature $\mathbf x$ and the output label $y$.\\
        $\mathcal D$ & Data generating distribution $(\mathbf x, y)\sim \mathcal D$.\\
        $n$ & The number of training samples $(\mathbf x_i, y_i), i=1,\ldots n.$ \\ 
        $S$ & The training set $(\mathbf x_i, y_i)\in S$.\\
        $\mathcal F$ & The hypothesis class.\\
        $f$ & The learning function, perhaps a deep learning model, from the hypothesis class $f\in\mathcal F$.\\
        $\mathbf w$ & The weights of the learning function $f_{\mathbf w}\in \mathcal F$ to be estimated from training data.\\
        $d$ & Model dimension: i) estimating dimension $d=\mathrm{dim}(\mathbf w)$, ii) effective dimension $d_\lambda$,\\ & iii) VC dimension $d_{\mathrm{VC}}$.\\
        $\Rf$ & Rademacher complexity. \\
        $\mathcal L(\mathbf w)$ & The estimation loss.\\
        $g$ & complexity measure. \\
        $R(\mathbf w)$ & The risk function $\mathbb E_{\mathcal D} \{\mathcal L(\mathbf w)\}$.\\
        $R^*$ & The optimal risk $R^*=\min R$\\
        $\mathbf g$ & The loss gradient ${\partial \mathcal L(\mathbf w) \over \partial \mathbf w}$\\
        $\mathbf m$ & The momentum that smooths  the gradient $\mathbf g$ linearly.\\
        $C(f)$ & Memory or computation constraints imposed on the learning function $f$.\\
        $\mathcal C$ & The hypothesis class $\mathcal F$ constrained by $C(\cdot).$\\ 
        $\eta$ & The learning rate in SGD update $\mathbf w_{k+1} = \mathbf w_k - \eta 
        \mathbf g_k$.\\
        $Q(\cdot)$ & Quantization operator as a projection to lower bits.\\
        $p(\cdot)$ & The probability mass or the density function.\\
        $\sigma(\cdot) $ & Nonlinear activation function.
    \end{tabular}
    \caption{Mathematical notation and their brief description}
    \label{tab:notation}
\end{table}

Basic learning theory deals with the predictor function $f\in\mathcal F$, where $\mathcal F$ is called the \emph{hypothesis class} and $f$ is the machine learning model such as multi-layer perceptron, perhaps indexed by some continuous parameters $\mathbf w,$ say $ f_{\mathbf w}$. One may augment $\mathbf w$ with a set of discrete parameters such as the number of layers and the number of units per layer, to generalize weight estimation towards neural architecture search.
It makes sense to consider the cardinality of the class to be finite in practice  $ | \mathcal F |<\infty$ because any finite-precision function $f_{\mathbf w}$ (e.g. in 32 bit single precision) provides many but finite set of choices for $\mathbf w$. 
The parameter $\mathbf w$ is typically trained using optimization methods such as the stochastic gradient descent (SGD). Suppose $(\mathbf x, y) \sim \mathcal D$ denote observed data generated from distribution $\mathcal D$, where $x\in\mathbb R^d$ is the input feature and $y$ the output label. The goal is to find the function $f$ such that $f(\mathbf x)$ approximates $y$ well, i.e. to learn function $f$ from the training data $(\mathbf x, y)$. 
In other words we aim at finding the best model from the hypothesis class $\mathcal F$ according to the expected loss $\mathbb E (\mathcal L\{\hat f(\mathbf x),y\}=R(\hat f)$ where the expectation is taken over the generating distribution $\mathcal D$, so $R(\cdot)$ is the true risk. In practice the empirical risk ${1\over n} \sum_{i=1}^n \mathcal L_{\mathbf w} (\mathbf x_i, y_i)$ is evaluated and minimized. SGD is commonly used to optimize the empirical risk for deep models. Table~\ref{tab:mlcomplex} lists the training and inference complexity for a few well-known machine learning models.

There are special cases of SGD update that are commonly used in practice. Suppose the positive real number $\eta_k$ is the learning rate at iteration $k$. The common SGD updates the weights according to 
\begin{equation}
    \mathbf w_{k+1} = \mathbf w_k - \eta_k \mathbf g_k, 
\end{equation}
where $\mathbf g_k$ is the gradient $\partial \mathcal L \over \partial \mathbf w_k$. The SGD with momentum updates the weights according to 
\begin{equation}
  \mathbf w_{k+1} = \mathbf w_k - \eta_k \mathbf m_k, 
\end{equation}
where $\mathbf m_k = \beta_1 \mathbf m_{k-1} + (1-\beta_1 ) \mathbf g_k$ is the momentum. These updates are often implemented in 32 bit float, but AI industry is pushing these  computations in lower bits; e.g. Google'e brain float that uses 16 bits, or the Grace Hopper NVIDIA chip that uses 8-bit float.


\begin{table}[t]
    \centering
    \begin{tabular}{c c c c  }
    Learning Algorithm     &  Model Size & Training Complexity & Inference Complexity  \\
    \hline
    Decision tree & $\mathcal O (n)$ & $\mathcal O (nd \log n)$ & $\mathcal O (\log n)$ \\
    Logistic regression & $\mathcal O(d)$ & $\mathcal O (nd^2+d^3)$ & $\mathcal O(d)$\\
    Multi-layer perceptron & $\mathcal O(dml)$ &  $\mathcal O(dmlnk)$ & $\mathcal O (dml)$  \\
    \end{tabular}
    \caption{Training and inference complexity of some common models; $d$ is the input dimension, $n$ is the training sample, $l$ is the number of layers, $m$ is the width of each layer, and $k$ is the number of epochs.}
    \label{tab:mlcomplex}
\end{table}

Empirical risk minimization averages the loss over the data samples $(\mathbf x_i, y_i)\in S, i=1,\ldots, n$ instead  $\mathcal D$, which introduces \emph{approximation error} and \emph{estimation error}  as explained in the following.
One may re-write that $R(f)-R^*$ differently, where $R^*$ is the true minimum over all possible functions. Note that the optimum function may probably fall out of the hypothesis class $\mathcal F$. This inductive bias of constraining $f\in\mathcal F$ calls for the following approximation and estimation error decomposition,  
\[
R(f)-R^*= \left\{\min_{f\in\mathcal F} R(f)-R^*\right\} + \left\{  R(f) - \min_{f\in\mathcal F} R(f)\right\},
\]
where the first term is the approximation error, and the second term is the estimation error. The above decomposition  facilitates better understanding of  finding a model $f$ whose risk is reasonably close to $R^*$ in and out of $\mathcal F$.

\begin{figure}[t]
    \centering
    \includegraphics[width=0.9\textwidth]{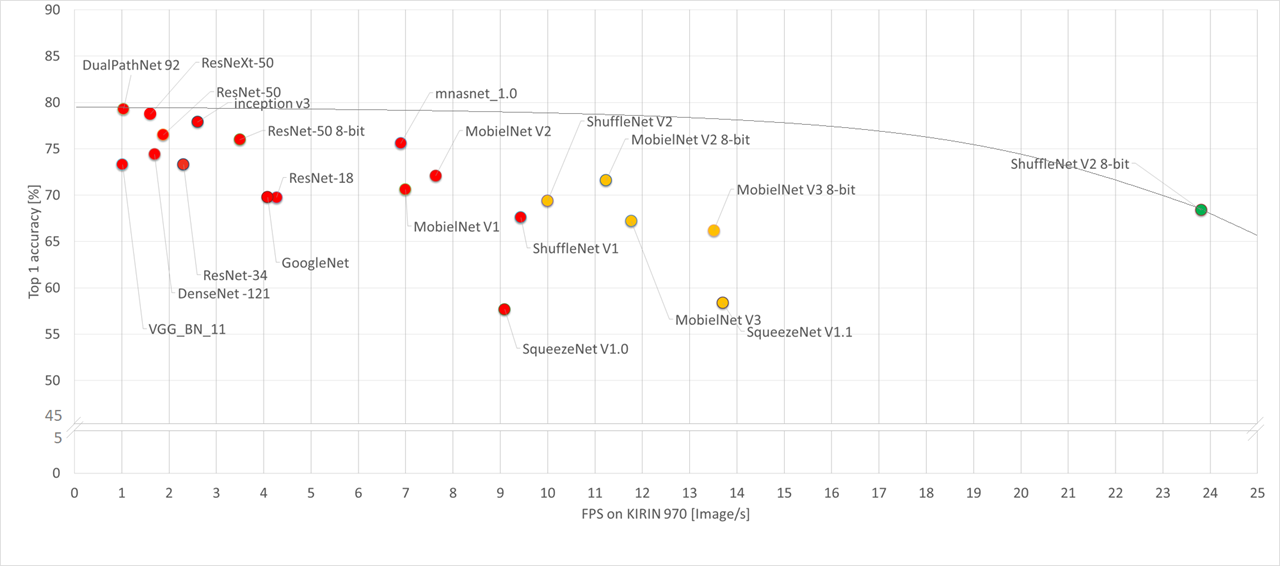}\\
    \includegraphics[width=0.9\textwidth]{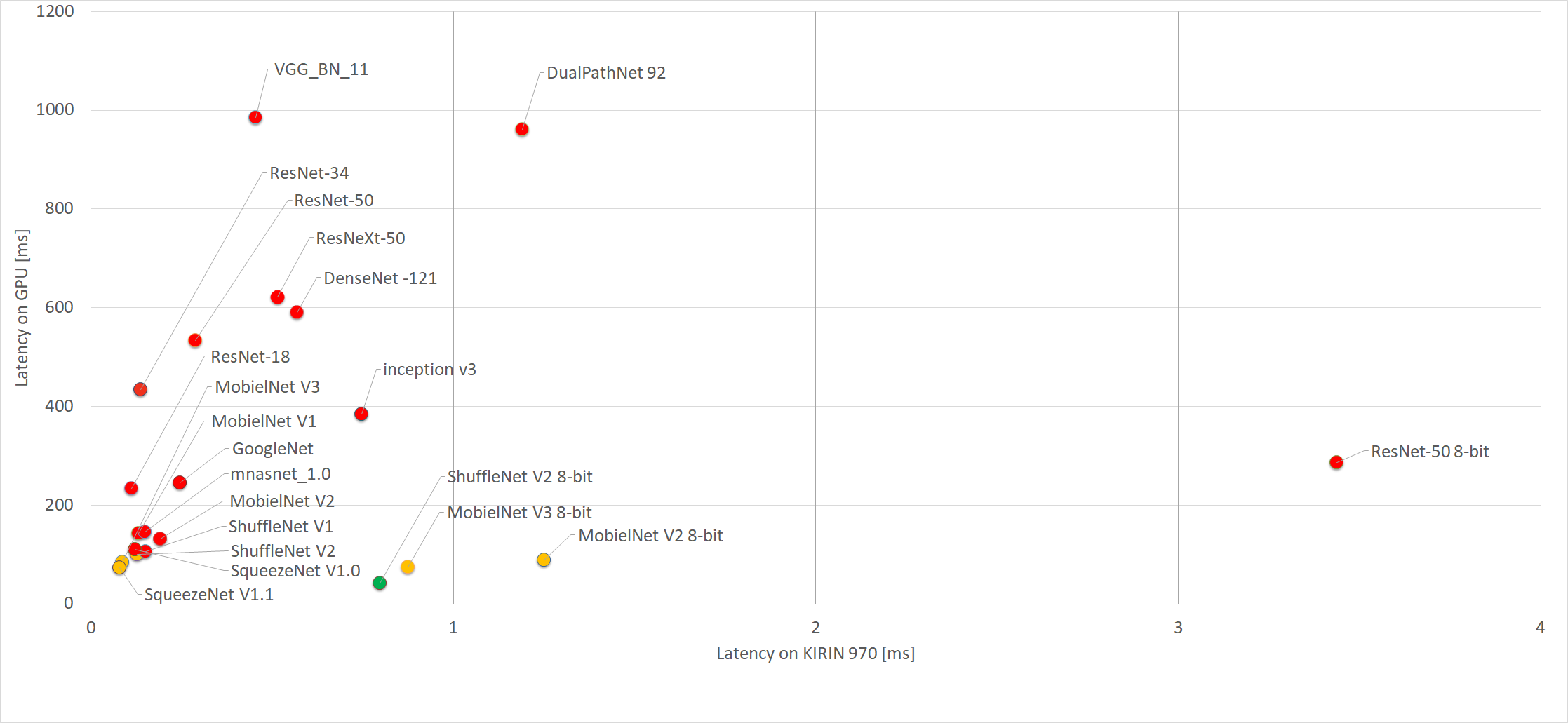}
    \caption{Famous ImageNet  classification models: full-precision versus  8bit quantized run on ARM CPU of Huawei Kirin 970  (top panel). Latency on NVIDIA V100 GPU versus latency on ARM CPU of Huawei Kirin 970, Latency on GPU cannot predict CPU deployment (bottom panel). }
    \label{fig:mit-accel}
\end{figure}

\section{Inference}

The main challenge of large models is to train and deploy them while the resource is constrained according to $C(\hat f)$ due to power, memory, and latency consumption. The common solution is to embed the computations in lower bits. Figure~\ref{fig:mit-accel} summarizes the state-of-the-art low bit solutions for deep models, see \cite{reuther2019survey} for a survey. 

More formally, we want to estimate the function $\hat f: \mathcal X \to \mathcal Y$ that minimizes $R(\cdot)$ while satisfying  $C(\hat f).$
The most important constraints are typically, i) memory, ii) latency, iii) energy. Most of the literature focuses on memory because it is difficult to model the latency  and power constraints as they are hardware dependent. In many scenarios, latency constraints can be translated into memory constraints for a given hardware.

Memory constraints appear at inference to fit the model into registers. For instance, the deep model $\hat f_{\mathbf w}$ is indexed with weights $\mathbf w$ and the weight value $\mathbf w$ has a certain range like $\pm 3.4\times 10^{38}$ if $\mathbf w$  is 32-bit float,  and $\mathbf w \in \{0,\ldots, \pm 2^{15}\}$ if the model is 16 bit integer. The range and the resolution of computation define the memory capacity.
\subsection{Learnability}
Learnability of a class is perhaps one of the most crucial properties required to ensure appropriateness of the chosen loss and class.  In a learning problem the true risk is minimized, i.e. $\hat f = \argmin_{f\in\mathcal F} R(f)$, and the optimum risk within the class is $ R(\hat f)\geq R^*$. In practice, however, the empirical risk is minimized, i.e. $\hat f_n = \argmin_{f\in\mathcal F} {1\over n} \sum_{i=1}^n\mathcal L\{f(\mathbf x_i, y_i)\}$. A class is learnable if  the risk of $\hat f_n$ approaches to $\hat f$, 
\[
\lim_{n\to \infty} \Pr\{R(\hat f_n) - R(\hat f) > \eps\} =0.
\]
This convergence must be uniform on the probability distribution $\mathcal D$ so that a class becomes a \emph{learnable class}. If  a uniformly convergent sequence of $\hat f_n$ does not exist, the class is not learnable. If such a sequence exists, the rate of convergence of $R(\hat f_n) \to R(\hat f)$  defines how hard it is to learn from data. For instance, decision trees are hard to learn  because this convergence rate is slow.
Suppose the constrained class is $\mathcal C = \{f\mid C(f)<c\}$. Before deploying the model at inference in low bits, one may need to make sure the low-bit version is learnable.
In other words, the constraint $C(\cdot)$ does not restrict the learnability of the class $\mathcal F$. In more precise terms $\mathcal F' = \mathcal F \cap \mathcal C$ is still a learnable class. If a class is learnable, the \emph{quantized} low-bit float or fixed-point projection of the class, $Q(\mathcal F)$, that reflects $\mathcal F'$ may or may not remain learnable.

\subsection{Lowbit Large Models}
In large deep models such as transformers \citep{vaswani2017attention}, the predictive function $f_{\mathbf w}$ even after training $\mathbf w$,  requires massive deployment resources. Suppose a model is already trained with weights $\hat{\mathbf w}.$ A common deployment strategy is to look for a low bit projection $Q(f_{\hat{\mathbf w}})$.The first step is to quantize the weights $Q(\hat{\mathbf w})$, and the second step is to implement the internal computations of $f_{\hat{\mathbf w}}$ in low bits. There are three strategies to look for a lower-bit projection: i) a data-free method in which only the model is used, ii) only a small calibration set of data is used iii) the whole training data is used. Methods i) and ii) are referred to as \emph{post-training quantization} while  iii) is called \emph{quantize-aware training}. With the advent of large models i) and ii) attracts more attention. The quantize-aware training is recently dismissed because the training data is often unavailable, and also the resources required for retraining a smaller model are very costly. Quantizing weights only, can be re-written simply as  
\begin{equation}
 Q(\hat{\mathbf w}) = \argmin_{\mathbf w\in Q(\mathbb R^d)} \lVert \mathbf w-   \hat{\mathbf w} \rVert,
\end{equation}
where the optimization is performed on the discrete set $Q(\mathbb R^d)$. A common method is  to choose a good $Q(\hat{\mathbf w})$ directly, for instance, a step function that transforms a continous $\hat{\mathbf w}$ to a discrete $Q(\hat{\mathbf w})$, which is known as the \emph{quantization function}. However, quantizing weights directly using the quantization function, does not assure a good approximation of $\hat{\mathbf w}$, because a deep model is composed of several layers and the approximation error of each layer affects the computation of the next layer. Ignoring the inter-layer computations by focusing on $\lVert \hat{\mathbf w}-\mathbf w \rVert$ may lead to a large function approximation error $\lVert f_{\hat{\mathbf w}}-f_{\mathbf w} \rVert$.   
Perhaps a better strategy is to choose the weights so that the output of the function is properly approximated, i.e.
\begin{equation}
 Q(\hat{\mathbf w}) = \argmin_{\mathbf w\in Q(\mathbb R^d)} \lVert f_{\mathbf w}-   f_{\hat{\mathbf w}} \rVert,
 \label{eq:quantout}
\end{equation}
but the tedious computation of $f$ makes this optimization infeasible. A greedy approach is used to match each layer instead.
A deep model is composed of several layers starting with the input features $\mathbf X_0$, which is a matrix of dimension $n\mathrm{dim}(\mathbf x)$, built by concatenating the input features $\mathbf x_i$. Each layer $l$ includes a weight matrix $\mathbf W_l$. The collection of such matrices forms the total weight $\mathbf w$, such that $\mathbf X_l = \sigma(\mathbf X_{l-1} \mathbf W_{l-1})$ where $\sigma(\cdot)$ is the nonlinear activation function. In each layer the following optimization is performed
\begin{equation}
    Q(\hat{\mathbf W_l})= \argmin_{\mathbf W_l}\lVert \sigma(\mathbf X_l\mathbf W_l) - \sigma(\mathbf X_l\hat{\mathbf W_l}) \rVert.
    \label{eq:blockquant}
\end{equation}
Quantizing the weights of layer $l-1$ will affect quantizion of the next layer $l$, and of course, calibration data are required to feed $\mathbf X_0$. A more precise quantization can be performed by optimizing \eqref{eq:quantout} directly, or by priortizing a  block of leading layers in the approximation error in \eqref{eq:blockquant}.

\section{Training}
Suppose the hypothesis class $\mathcal F$ is learnable. The challenge is to devise a computationally efficient algorithm $A(\cdot):S\to\mathcal F$ that uses the training data $(\mathbf x_i, y_i) \in S$ to pave the way towards finding a good candidate function $\hat f\in\mathcal F$. This is equivalent to finding $f_{\mathbf w}$, i.e. estimating $\hat{\mathbf w}$. In deep learning the number of parameters $d=\mathrm{dim}(\mathbf w)$ is overwhelming. A common remedy for large resource requirements is to lower the number of bits, from the standard 32 bit single-precision float towards 16 bit half-precision, or even lower \citep{hubara2017quantized}. The loss of model accuracy is the main obstacle in lowering the number of bits. Training on 8 bits \cite{ghaffari2022integer}, and inference on 8 bits \cite{wu2020integer} would not hurt the accuracy compared to single precision in practice. However, the limits of lowering the bit width with no accuracy loss is still being evaluated empirically, and require more theoretical study \citep{cacciola2023convergence, metel2022variants, zhang2022low}.

The training constraint is two-fold \citep{steinhardt2016memory}, i) the constraint on the hypothesis class $\mathcal F$, ii)  the constraint on the approximating algorithm $A(\cdot):S\to\mathcal F$ towards estimating $\hat f \in \mathcal F$. 
Suppose $A(S)$ leads to a proper $\hat f$, given $\mathcal F\cap \mathcal C$ is learnable. 
The main challenge is to find a training algorithm $A:S\to\mathcal F$ that minimizes $R(\cdot)$ while it satisfies $C\{A(S)\}$ to deliver $Q(\hat f)$. In the sequel we only focus on the common training algorithms for deep models, i.e. we assume $A(\cdot)$ to be a low-bit SGD.

\subsection{Lowbit SGD}
Neural network training has been performed in single-precision (32-bit) floating-point. The ever-increasing size of deep learning models motivated the use of lower precision data types, such as low-bit floating, fixed, or dynamic fixed-point number representations during model training and for the final model representation. Besides decreasing memory requirements, model training and inference time can be reduced, as well as hardware and electricity costs. This makes  the development and the use of deep models accessible to more people.

A large body of research uses different number formats for different types of data to save resources while at the same time maintaining the model accuracy achieved using single precision, see Table \ref{t:1}. Model weights are quantized during or after training. Often only the most time-consuming operations, such as matrix multiplication are performed in a low-bit format. Given the difficulty in training neural networks, certain non-linear operations or weights are typically left in full precision.  

\begin{table}
	\caption{Taken from \citep[Table 1]{wang2022}, this table shows the low-bit integer formats used in each paper for the weights ($w$), weight accumulators ($w$ acc), activation functions ($a$), weight gradients ($g$), activation gradients ($e$), and softmax, where fp32 denotes single-precision floating-point.}	
	\label{t:1}
	\begin{center}		
		\begin{tabu}{l|cccccc}			
			\hline
			&$w$&$w$ acc&$a$&$g$&$e$&softmax\\
			\hline
			\citep{zhu2017}&2&32&32&32&32&fp32\\
			\citep{rastegari2016}&1&32&1&32&32&fp32\\
			\citep{courbariaux2015}&1&32&32&32&32&fp32\\
			\citep{jacob2018}&8&32&8&32&32&fp32\\
			\citep{zhou2016}&1&32&2&32&6&fp32\\
			\citep{banner2018}&8&32&8&32&8&fp32\\
			\citep{wu2018}&2&8&8&8&8&fp32\\
			\citep{chen2017}&1&12&1&12&12&fp32\\
			\citep{das2018}&16&32&16&16&16&fp32\\
			\citep{wang2022}&8&8&8&5&8&integer\\
			\hline
		\end{tabu}
	\end{center}
\end{table} 

An existing gap between optimization theory and neural network training is amplified by the use of low-bit number formats. Almost all optimization theory is developed in Euclidean space, with its convergence results relying on concepts such as continuity, limits of sequences, gradients, etc., whereas neural network training is performed numerically in finite precision environments. Unlike single-precision floating-point, the gap between theory and computation cannot be ignored in general given non-trivial rounding errors in low bits.

\begin{figure}
	\centering
	\begin{tikzpicture}[xscale=3, yscale=1.5]
		\draw [ultra thick, <->] (0.5,1) -- (0,0) -- (1,0.5);
		\node [above left] at (0.5,1) {$\mathbf g_k$};
		\node [right] at (1,0.5) {$\mathbf g_k+\mathbf e_{1k}$};
		\draw [cyan,thick, -] (0.5,1.0) -- (1.0,0.5);
		\node [cyan, right] at (0.7,0.95) {$\mathbf e_{1k}$};
	\end{tikzpicture}
	\caption{The vector $\mathbf g_k$ is the desired stochastic gradient, but $\mathbf g_k+\mathbf e_{1k}$ is the resulting computed stochastic gradient due to the computational error $\mathbf e_{1k}$ from low-precision computation.}
	\label{f:1}
\end{figure}
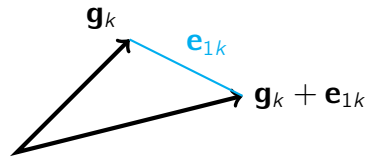

A step of SGD can be modelled as 
\begin{alignat}{6}
	\mathbf w_{k+1}=\mathbf w_k-\eta_k(\mathbf g_k+\mathbf e_{1k})+\mathbf e_{2k},
 \label{eq:errorsgd}
\end{alignat}
where $\mathbf w_k$ are the trainable parameters of the neural network in iteration $k$, $\eta_k$ is the step-size, $\mathbf g_k$ is a stochastic gradient, $\mathbf e_{1k}$ is the rounding error from approximately computing $\mathbf g_k$, see Figure \ref{f:1}, and $\mathbf e_{2k}$ is the rounding error from computing all of the arithmetic operations in \eqref{eq:errorsgd}. The error $\mathbf e_{1k}$ is most problematic given that all arithmetic operations in computing the forward and back propagation contribute to it, increasing its upper bound as the model size increases. In addition, unbiased error $\E(\mathbf 
 e_{1k})=\mathbf 0$ does not hold in general, even when using stochastic rounding. 
The convergence of gradient descent with computational error in the gradient is a long-studied problem, see for example \citep[Chapter 4]{polyak1987} and \citep[Chapter 1.2]{bertsekas1999}. Recently  \cite{xia2022,cacciola2023convergence,metel2022variants}, studied the convergence of gradient descent in low-precision environments. Taking all computations to a sufficiently low precision will destroy the error assumptions in these works, implying the inability of \eqref{eq:errorsgd} to converge in general.

We wonder if stochastic gradient descent is still a viable training algorithm for general low-precision neural network training. Especially if the low-precision number format no longer sufficiently approximates Euclidean space. Perhaps a more appropriate algorithm should be used, acknowledging that the optimization is being performed in a finite space. The error $\mathbf e_{1k}$ could be decreased by using a finite difference approach to approximately compute the gradient. An alternative is to abandon approximate gradient methods for purely heuristic search methods used for black-box optimization adopted for low bit training structures.

\subsection{Effective Parameters}
\begin{figure}
    \centering
    \includegraphics[width=0.5\textwidth]{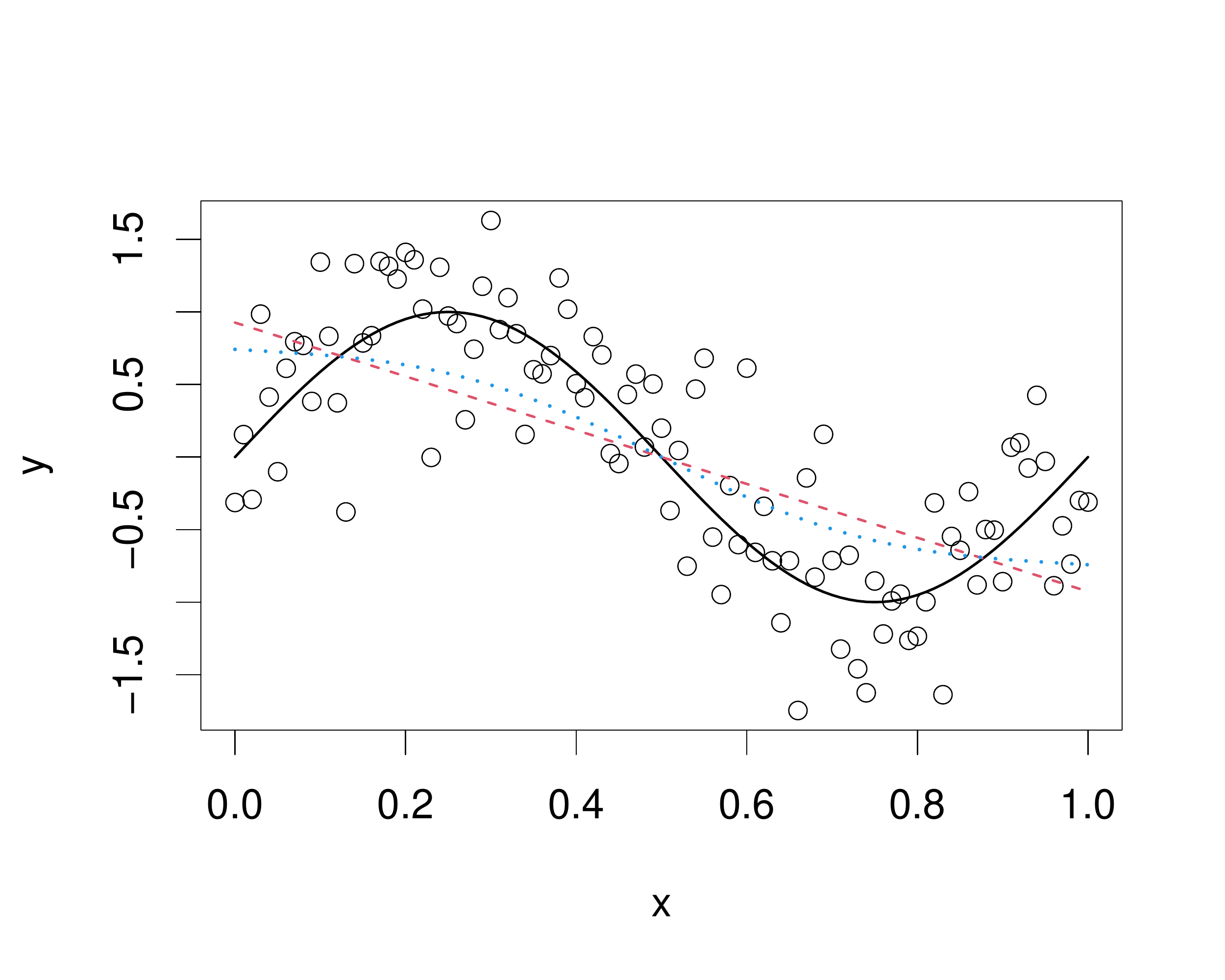}
    \caption{Effective parameters of cubic local polynomial with 100 parameters, but the effective parameters is tuned by $\ell_2$ regularization to be equal 2 (red dashed), equal 3 (blue dotted), and equal 5 (solid black). The lottery ticket hypothesis is more likely to be true if effective parameters and model parameter differ significantly. }
    \label{fig:smooth}
\end{figure}

Deep learning models include many parameters that overloads their computation. This complicates the training, because all such large models require proper and mostly complicated regularization schemes. Redesigning a smaller model trained with a lower amount of regularization can not only simplifies training, but also lead to lower resource inference. This requires rethinking the regularization concept, and calls for a new optimization algorithm that relates the large and highly regularized models to smaller and less  regularized models. The concept of effective parameters allows us to have an idea about a new model that can approximate the original model with good accuracy but smaller number of parameters. This is closely related to the complexity of the true underlying model. Figure~\ref{fig:smooth} illustrates how the number of parameters decreases as more regularization is exercised in training.

\cite{ye1998measuring} formalizes the linear model fit with $\ell_2$ regularization 
$$\hat{\mathbf w_\lambda} = \argmin_{\mathbf w\in \mathbb R^d} \lVert \mathbf y-\mathbf X \mathbf w\rVert +\lambda \lVert \mathbf w \rVert$$
$$d_\lambda:= \mathrm {dim}(\hat{\mathbf w}_\lambda) \propto {\mathrm{cov}(\mathbf y, \hat {\mathbf y_\lambda}) },$$ 
where  $\hat y_\lambda  = \mathbf X \mathbf w_\lambda$, and $d_\lambda$ decreases as $\lambda$ increases \citep{efron2004estimation}. 
 This concept is closely related to compression bound \citep{blier2018description}, geometric complexity \citep{dherin2022neural}, and generalization error \citep{ji1993generalization,grant2022predicting}. 

We wonder how to extend this concept to deep models to have an idea about their effective dimensions. Even knowing the effective parameters may not help to construct the smaller model. A proper algorithm to find a more compact model given the effective number of parameters is still an open research question.

\subsection{Data Dimension}
Data used in deep learning such as image pixels, language words, or speech intensity has a  low-dimensional structure despite the high-dimensional representation. This property is the reason for the remarkable success of deep models. The common intuition is that each layer folds the dimension through a nonlinear activation before and passes the folded information to the next layer. The data dimension is model-free and only relies on data only. \cite{levina2004maximum} suggests to count the neighbouring points  to estimate the data dimension and \cite{pope2021intrinsic} shows the impact of data dimension on learning.

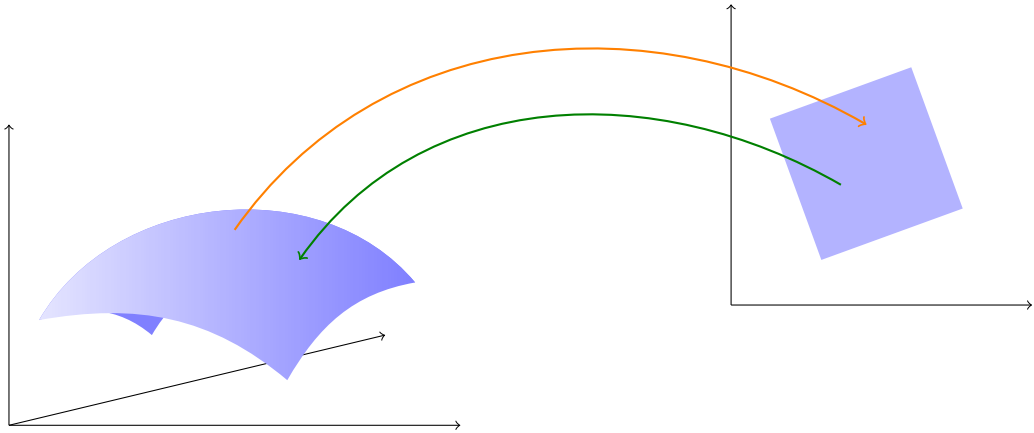
\begin{figure}
\begin{tikzpicture}[scale=2]
\draw[->] (0, 0) -- ++(0, 2);
\draw[->] (0, 0) -- ++(2.5, 0.6);
\draw[->] (0, 0) -- ++(3, 0) node[midway, below, yshift=-0.5em]
    {};

\draw[fill=blue!50, draw=none, shift={(0.2, 0.7)},scale=0.5]
  (0, 0) to[out=20, in=140] (1.5, -0.2) to [out=60, in=160]
  (5, 0.5) to[out=130, in=60]
  cycle;

\shade[thin, left color=blue!10, right color=blue!50, draw=none,
  shift={(0.2, 0.7)},scale=0.5]
  (0, 0) to[out=10, in=140] (3.3, -0.8) to [out=60, in=190] (5, 0.5)
    to[out=130, in=60] cycle;

  \draw[->] (4.8, 0.8) -- ++(0, 2);
  \draw[->] (4.8, 0.8) -- ++(2, 0) node[midway, below, yshift=-0.5em]
      {};

  \draw[thin, fill=blue!30, draw=none, shift={(5.4, 1.1)}, rotate=20]
    (0, 0) -- (1, 0) -- (1, 1) -- (0, 1) -- cycle;

  \draw[thick,->,orange]
    (1.5, 1.3) to [out=55, in=150] node[midway, above, xshift=6pt, yshift=2pt]
    {} (5.7, 2);

  \draw[thick,->,green] (1.5, 1.3) ++(4.03, 0.3) to [out=150, in=55]
    node[midway, below, xshift=2pt, yshift=-2pt] {} ++(-3.6, -0.5);

\end{tikzpicture}

    \caption{ A three-dimensional manifold with the intrinsic dimension of two.}
    \label{fig:lowdim}
\end{figure}

Given a set of sample points in $\mathbb R^n$, it is common to assume that the data lies on or near a low-dimensional manifold, see Figure~\ref{fig:lowdim}. The common approach is to use a Poisson process to model the number of points found by random sampling within a given radius around each sample point. By relating the rate of this process to the
surface area of the sphere, the likelihood equations yield an estimate of the inverse \emph{intrinsic dimension} at a given point $\hat d^{-1}(\mathbf x_i)$. Therefore ultimate estimation $\hat d^{-1}$ is averaging $\hat d^{-1}(\mathbf x)$ over the $n$ data points to provide an estimation of the data dimension  \citep{mackay2005comments}
\begin{equation}
    \hat d \approx\left\{ {1\over n} \sum_{i=1}^n \hat d^{-1}(\mathbf x_i)\right\}^{-1}.
    \label{eq:datadim}
\end{equation}

This means each data point carries a weight about the true data dimension.  Intuitively weighting samples leads to weighting their respective fitted models \citep{friedman2000additive}. In other words models are smoother version of data. Therefore, an alternative data dimension estimation  can be deployed through the concept of \emph{effective parameters} explained earlier. While models vary in parameter size, their effective dimensions remain close to the true data dimension. One may call for  an algorithm that  estimates data dimension during training by connecting effective parameters with batch data dimension to lower the parameters of the model and compress while training, simultaneously.  An ideal model uses the training parameters effectively and matches the data dimension with the model dimension. An ideal  descent direction takes the gradient in two direction i) weight direction ii) model size direction. In training step,  optimal weights are found given the dimension, and in compression optimal dimension is found given the weights. 

\section{Ambient and Intrinsic Dimension}

It is widely believed that deep neural networks work well when data are essentially on a low-dimensional manifold embedded in a high-dimensional ambient space. This view is particularly pervasive for natural image data \cite{pope2021intrinsic} and \cite{ansuini2019intrinsic}. \cite{shaham2018provable} formalize this view by proving, under some conditions, that the universal approximation depends strongly on the intrinsic dimension of the data while the dependence on the dimension of ambient space is comparatively rather weak. The knowledge about the intrinsic dimension is therefore imperative in order to decide how well neural networks work and how they should be designed. The more recent work by \cite{imaizumi2019deep} and \cite{imaizumi2022advantage} formally establish the advantage of DNN in estimating non-smooth functions. \cite{nakada2020adaptive} further show that the optimal minimax rate is achievable using DNN and the rate essentially depends on the intrinsic dimension measured using Minkowski's fractal dimension. More recent studies by \cite{birdal2021intrinsic} sheds further light on computational aspects of intrinsic dimension and connection to generalization of DNN. It is, however, remain to understand how intrinsic dimension is related to the depth and width of DNN. To be more concrete given the intrinsic dimension of data, what is the minimal depth and width to achieve a pre-specified level of accuracy in training and latency in inference.  

While different deterministic approaches for measuring intrinsic dimension using variants of Hausdorff (topological) dimension, including Minkowski's, persistent homology based measures of dimension or other methods aim at measuring the dimension of the whole data cloud, one may take a statistical perspective and try to measure the dimension of a manifold that can cover the great majority, say over 90 or 95 percent, of the data cloud. In view of the concentration phenomena in large dimension, it is plausible to expect such  approach leads to a much smaller intrinsic dimension. To establish what that have been already studied by the aforementioned authors using such statistical approaches in measuring the intrinsic dimension seems a fruitful direction in studying advantages of deep neural networks. Further to such studies, one may try to answer the question posed in the previous paragraph about the connection between data dimension and the hyper-parameters of deep neural networks. 
A thorough study on effective methods of dimension estimation and universal approximation of deep neural networks can hopefully lead to an explicit, though approximate, formula connecting the intrinsic data dimension to the architecture of deep neural networks. Such studies can provide guidelines for at least part of neural network architectures.


\section{Optimizer}
\label{opt_nn}
Different modifications of stochastic gradient descent (SGD) have been successfully used for the optimization (training) of neural networks. The method constitutes the iterative updates of model weights ideally reaching a lower value of loss function at each step. 

A variety of learning rate schedulers (dependency of the learning rate $\eta_k$ on $k$, the iteration step) are used in practice. The simplest one is the constant learning rate (all $\eta_k$ are the same). Usually practitioners apply some kind of decay on the learning rate during training ($\eta_k$ is a monotone decreasing function of $k$). \cite{goyal2017} showed the importance of learning rate warm-up for some settings: starting with a very small $\eta$, then increase it during the training and then anneal back. 

There are also implicit ways to modify the learning rate, so called  
adaptive methods.
\begin{itemize}
\item RMSProp: 
\begin{equation}
    \mathbf w_{k+1} = \mathbf w_k - \frac{\eta}{\sqrt{\mathbf v_k}} \mathbf g_k, \ \text{where} \ \mathbf v_k = \beta_2 \mathbf v_{k-1} + (1-\beta_2 ) \mathbf  g_k^2 \ \text{is the second momentum.}
    \label{eq:update_rmsprop}
\end{equation}
\item Adam \citep{adam}: 
\begin{equation}
\begin{split}
    \mathbf w_{k+1} = \mathbf w_k - \frac{\eta}{\sqrt{\hat{\mathbf v}_k}} \hat{\mathbf m}_k, \ \text{where} \ \hat{\mathbf m}_k = \frac{\mathbf m_k}{1-\beta^k_{1}}, \ \hat{\mathbf v}_k = \frac{\mathbf v_k}{1-\beta^k_{2}} \\
    \text{are unbiased estimators of the first and second momentums.}
    \end{split}
    \label{eq:update_adam}
\end{equation}
\item LARS \citep{lars}: 
\begin{equation}
    \mathbf w_{(l),k+1} = \mathbf w_{(l),k} - \eta\frac{\|\mathbf w_{(l),k}\|}{\|\mathbf g_{(l),k}\|} \mathbf g_{(l),k}, \ \text{where} \ (l) \ \text{corresponds to the $l$-th layer parameters.}
\label{eq:udpate_lars}
\end{equation}
\end{itemize}

It has been formally proven that adding momentum gives an acceleration in convergence for stochastic gradient methods \citep{polyak, nesterov, Danilova}. 
Adaptive optimizers are not guaranteed to converge to the optimal solutions even in the convex case \citep{reddi2018}, but in practice, it has been demonstrated to be fast and reliable \citep{adam}. SGD with momentum could outperform adaptive optimizers in vision tasks \citep{Keskar2017ImprovingGP}, however, adaptive methods become especially important for attention models like transformers \citep{ZhangKVKRKS20}. Finally, adaptive optimizers of LARS type help stabilizing the training with large batch sizes and hence increasing the training speed \citep{lamb}. 

A general optimizer update rule  can be written as 
\begin{equation}
   \mathbf w_{k+1} = h(\mathbf w_{1:k}, \mathbf  g_{1:k}, k, \mathcal F)
   \label{eq:gen_update}
\end{equation}
 There has been an effort \citep{Andrychowicz2016LearningTL, velo2022} to learn the update function $h$ from \eqref{eq:gen_update} in a meta-learning setting for different task and architectures. Although being promising, this approach still doesn't scale well to large network and requires expensive training.



 The main problem with tracing first and second-order momentum is the memory. One needs roughly $3\times$ more memory for the gradient update, which becomes especially problematic with training large neural networks. Furthermore, SGD  provides the solution with better generalization than its more advanced counterparts like ADAM  \citep{wilson2017}. However, utilization of momentum stabilizes the training  and in some cases accelerates it. Momentum is currently applied in most optimizers for large networks. Is utilizing momentum really necessary for training and is there a way to achieve stability and acceleration without them? 

Ideally we want to modify  history dependency in the update equation, while keeping the training stable and efficient. It would be highly beneficial for practitioners to find the effective version of SGD applicable to many deep model, in particular, large transformers. We wonder if there is a way to design network-specific optimizers rather than using default methods with cumbersome hyperparameter fine-tuning. 
In other words, we wonder how to utilize the inductive bias $f_{\mathbf w}\in \mathcal F$ \citep{goyal2022inductive} to design an effective (and efficient!) update step with some theoretical guarantees on convergence.

\section{Generalization}
One of the most profound and broadest math challenges in deep learning is the generalization problem. For example, in autonomous driving, the training environments (e.g.~daytime in a park) often differ from the test environments (e.g.~night in an urban area); the training text in a sentiment analysis system differs from the real text to classify. The generalization problem deals with  obtaining a machine learning model with good performances on our training datasets, that  formally guarantee it also performs well on new datasets. There are two types of assumptions for this problem i) In-domain (ID) generalization, i.e.~samples from the training set and test set are both drawn from the same underlying distribution and ii) out-of-domain (OOD) generalization in which the training set and the test set are drawn from different underlying distributions. For classification, a sample $(\xv, y) \sim \Dc$ is composed of input $\mathbf x$ (e.g.~an image) and label output $y$. 

In this section, we summarize existing mathematical formulations of both in-domain and out-of-domain generalizations. We denote $\Dc$ as an underlying distribution (domain) and $S$ as a finite set of $n$ samples from $\Dc$. We use $\mathcal F$ as a shorthand of the hypothesis 
 class and $f$ as a hypothesis (model). We focus on the classification task throughout, but generalization bounds for other tasks (such as regression) are also possible \citep{mohri2018foundations}. In classification, a sample $(\xv, y)\sim \Dc$ is composed of input $\xv$ (e.g.~an image) and output $y$ (a label). 

\subsection{In Domain}\label{eq:in-domain}

To evaluate the performances, we need to define the evaluation metric. In classification, the default choice is classification error, i.e., the percentage of wrong predictions. Given a model $h$, the error on domain $\Dc$ and on the sample set $S$ are computed as the following:
\begin{align}
\epsilon_\Dc(f) = \Eb_{\Dc} [\one\{f(\xv)\neq y\}], \,\quad \epsilon_S(f) = \Eb_{S} [\one\left\{f(\xv)\neq y\right\}],
\end{align}
where $\one(\cdot)$ is an indicator function. The goal of the in-domain generalization is to provide the following bound
\begin{align}\label{eq:in-domain-bound}
\epsilon_\Dc(f) \leq \,\epsilon_S(f) + g(n, \mathcal F),
\end{align}
and the function $g(n, \Hc)$ represents the \emph{generalization gap} between test and training errors. Ideally, we want $g(n, \Hc) \to 0$ as $n \to \infty$. The dependence on $\Hc$ is usually characterized by the model capacity, i.e., how expressive our model class is. In many scenarios, the number of samples we can collect is limited (e.g.~in healthcare). On the other hand, modern models contain millions or even billions of parameters (e.g.~Transformers, \citealt{vaswani2017attention}). Therefore, the exact form of $g(n, \Hc)$ will guide us towards
i) finding how many samples are necessary and this is important since labelling is costly in practice; ii) finding the right model architecture. Even though Transformers could contain billions of parameters, in many cases, they do not suffer from overfitting. This requires a better understanding of the function $g(n, \Hc)$, which could help us design better model architectures or even conduct model compression. 

Unfortunately, existing bounds of type \eqref{eq:in-domain-bound} are often vacuous for neural network models. 

For binary classification, the earliest model capacity measure is called Vapnik--Chervonekis (VC) dimension \citep{vapnik1971uniform, valiant1984theory, blumer1989learnability}. With VC dimension, we can obtain the uniform convergence bound \citep[e.g.][Corollary 3.9 and Theorem 3.17]{shalev2014understanding}:
\begin{align}
\Pr\left\{\epsilon_\Dc(h) \leq \epsilon_S(h) + \sqrt{\frac{2 \log \sum_{i=0}^{d_\mathrm{vc}} \binom{n}{i}}{n}} + \sqrt{\frac{\log(1/\delta)}{2n}}\right\} \geq 1 - \delta,
\label{eq:vc_dim}
\end{align}
where $d_\mathrm{vc}$ is the VC dimension. This theorem tells us given the training error, the VC dimension, and the number of i.i.d.~samples, we can provide an upper bound for the test error. For ReLU networks, a nearly tight VC dimension bound has been given in \citep{bartlett2019nearly}. 

For example, suppose our dataset is MNIST \citep{lecun1998gradient}, and there are $m = 50,000$ samples. Using Theorem 7 from \citet{bartlett2019nearly}, one can obtain that the VC dimension is around  $4.4 \times 10^6$ for a two-hidden layer MLP where each hidden layer has $256$ neurons. Plugging it back into \eqref{eq:vc_dim} we obtain:
\begin{align}
\epsilon_\Dc(f) \leq \epsilon_S(f) + 1.185.
\end{align}
Since both $\epsilon_S(f)$ and $\epsilon_\Dc(f)$ are between $0$ and $1$, this bound does not provide us with a vacuous guarantee. This problem is even worse for large models with billions of parameters.

Suppose we have $n$ i.i.d.~samples from a distribution $\Dc$ and a machine learning model $f$ from class $\Hc$ that can achieve good performance on the training set. Can we provide theoretical guarantees for the test performance of $f$ on $\Dc$ that could guide model selection and data collection?


\subsection{Complexity}
 An alternative model capacity measure is the Rademacher complexity \citep{koltchinskii2001rademacher}. Similar to \eqref{eq:vc_dim}, the Rademacher complexity bound \citep[e.g.][Theorem 3.5]{mohri2018foundations} can be written as
\begin{align}
\Pr\left\{ \epsilon_\Dc(f) \leq \epsilon_S(f) + \Rf_n(\Hc) + \sqrt{\frac{\log(1/\delta)}{2n}}\right\} \geq 1 - \delta.
\label{eq:rademacher}
\end{align}
The term $\Rf_n(\Hc)$ is called the Rademacher complexity. Intuitively, it is the capability of the function class $\Hc$ to fit random fair coins. Deriving tight Rademacher complexity is also a hot research topic in recent years. For example, \citet{neyshabur2018role} proposes a Rademacher complexity bound for two-layer ReLU networks that can partially explain the effect of overparametrization. For deep neural networks, a tight generalization bound is yet to be found. 

Other than the VC dimension and Rademacher complexity, there are other capacity measures that could potential explain generalization in deep learning, such as covering number \citep[e.g.][]{shalev2014understanding, zhu2021understanding}, PAC-Bayes bounds \citep{mcallester1998some, lotfipac}, compression schemes \citep{littlestone1986relating, ashtiani2018nearly}, and information theoretical bounds \citep{haghifam2021towards}. These generalization bounds have been applied to deep learning to partially explain the role of data augmentation, model size, model compression, etc. 

\begin{figure}
    \centering
    \includegraphics[width=0.8\textwidth]{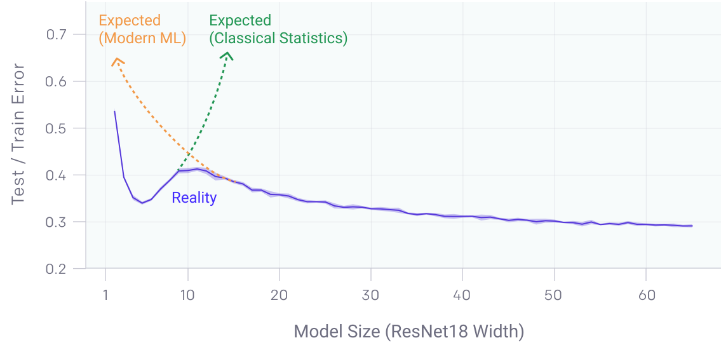}
    \caption{Illustration of the double descent phenomenon.}
    \label{fig:double_descent}
\end{figure}

The non-vacuous generalization bound can provide theoretical support for us to understand an important generalization phenomenon: double descent~\citep{belkin2019reconciling,nakkiran2021deep}: as illustrated in Figure~\ref{fig:double_descent}, as the model size increases, the performance of machine learning models first improves, then gets worse, and then improves again.

The modern double descent regime requires a new learning theory beyond the classical statistical one, e.g., VC dimension and Rademacher complexity mentioned above. We expect that a better approximated  $g(n, \Hc)$ could also help to explain the double descent phenomenon. This phenomenon is fairly universal that happens in CNNs, ResNets, transformers and even linear models~\citep{hastie2022surprises} as well as decision trees~\citep{wyner2017explaining}, and occurs in a wide variety of different tasks, including image classification and language translation. Therefore, solving the aforementioned mathematical challenge to provide the non-vacuous generalization guarantee is potentially beneficial for model and data size selection in real-world scenarios. In addition, this generalization phenomenon is also closely linked with optimization techniques we apply, such as SGD~\citep{keskar2016large,dinh2017sharp}. Involving the optimization analysis will be more mathematically challenging.



\subsection{Out of Domain}

Compared to in-domain generalization, a more challenging task is out-of-domain generalization. We have the distribution shift problem in this generalization, meaning that the training and test sets are drawn from different distributions. Let us assume that the data generating distribution is partitioned into $\mathcal D = \Sc \cup \Tc$ training set is from a source domain $\Sc$ and the test set is from a target domain $\Tc$. We also assume that each domain can be properly estimated to disentangle the out of domain generalization from the in-domain one.

If the target domain is not related to the source domain, then there is no hope that we can learn a model that performs well on the target. Therefore, there has to be some connection between the two domains. There are two popular types of domain shift under research:
\begin{itemize}
\item \emph{Covariate Shift: } the input distributions are different, i.e., $p_\Sc(\xv) \neq p_\Tc(\xv)$ for some $\xv$, but the conditional distributions are the same, i.e., $p_\Sc(y|\xv) = p_\Tc(y|\xv)$;
\item \emph{Lable Shift: } the label distributions are different, but the input distributions for each class are the same, i.e., $p_\Sc(y) \neq p_\Tc(y)$ for some $y$ but $p_\Sc(\xv|y) = p_\Tc(\xv|y)$ for all $(\xv, y)$.
\end{itemize}

First, we discuss the covariate shift case. The first generalization bound under the assumption of covariate shift is from \citet{ben2006analysis}. It gives an upper bound for the target (test) error based on the source (training) error:
\begin{align}\label{eq:cov}
\epsilon_\Tc(f) \leq \epsilon_\Sc(f) + \Delta_{\Hc}\left\{p_\Sc(\xv) , p_\Tc(\xv)\right\} + \lambda^*,\, \mbox{ for any }f\in \Hc.
\end{align}
The second term $\Delta_{\Hc}$ measures the difference between the two input distributions $\left\{p_\Sc(\xv) , p_\Tc(\xv)\right\}$, and $\lambda^* = \argmin_{f\in \Hc} \left\{\epsilon_\Sc(f) + \epsilon_\Tc(f)\right\}$ denotes the optimal joint error of the source and target domains. The term $ \Delta_{\Hc}\left\{p_\Sc(\xv) , p_\Tc(\xv)\right\}  + \lambda^*$ is the \emph{generalization gap}, as it is an upper bound of the gap between source and target domains. The generalization gap is small if i) the input distributions of $\Sc$ and $\Tc$ are close to each other; ii) the optimal joint error is small.

To achieve a small generalization gap, people use deep neural networks to embed the input distributions \citep{ganin2016domain, zhang2019bridging, acuna2021f}. These embeddings are called \emph{deep features}, and this method is called \emph{feature matching}. Suppose $g$ is the aforementioned neural network, feature matching requires $p_\Sc\{g(\xv)\} = p_\Tc\{g(\xv)\}$ for any $\xv \in \Xc$. Under the covariance shift assumption $p_\Sc\{y|g(\xv)\} = p_\Tc\{y|g(\xv)\}$, and thus the out of domain generalization vanishes while using $g(\xv)$ instead of $\xv$ and for instance the  Bayesian optimal classifier on both domains coincide. 


Second, we discuss the label shift scenario. Switching the roles of $\xv$ and $y$ in covariate shift, we obtain the label shift assumption. Under this assumption, \citet{zhao2019learning} argues that the optimal joint error in \eqref{eq:cov} can be lower bounded. Therefore, separate generalization bound under label shift is needed.  \citet{tachet2020domain} proposes a generalization bound based on the label shift $\Delta\{p_\Sc(y) , p_\Tc(y)\}$ and the domain shift of the conditional distribution $p_\Dc(\hat{y}|y)$, where $\hat{y}$ is the prediction. In order to minimize such generalization bound, \citet{tachet2020domain} aims to enforce the embedding-type approach and look for invariance of the class-conditional distributions $p_\Dc\{g(\xv)|y\}$. However, this will induce computational inefficiency when there are many classes.

In many cases, the aforementioned generalization bounds are difficult to verify in practice and similar to in-domain generalization, such bounds are often vacuous. We wonder if one can train a machine learning model on one or more source domains so that this model provably performs well on new target domains. We call to   define all such domains formally with easily verifiable criteria.

\section{Challenge}
In this section we aim at re-stating the challenges we explained in the text more concisely. 
\begin{itemize}
    \item [Lowbit model:] Given $\mathcal F$ is learnable, we wonder if  a lower bit projection $\mathrm P_\Omega (\mathcal F)$ where $\mathrm P$ is the projection function and $\Omega\subset\mathbb R$ is the space of the lower bit fixedpoint or floating point representation of $f\in \mathcal F$.
    \item [Constraint:] Given $\mathcal F$ is learnable, we wonder a constrained version $\mathcal F \cap \mathcal C$ is also learnable. The constraints may reflect memory (hardware independent), or latency (hardware dependent).
    \item [Quantization:] Quantizing large language models allows to run on the lower resource cloud and edge. We wonder how one can optimize \eqref{eq:quantout} more effectively. For instance instead of optimizing  $\lVert \sigma(\mathbf X_l\mathbf W_l) - \sigma(\mathbf X_l\hat{\mathbf W_l}) \rVert$ over layers $l$ as in \eqref{eq:blockquant} we may look for a model subgraph $\mathcal G$ and alternate between optimizing $\min_{\mathbf W \mid \mathcal G}$ and $\min_{\mathcal G\mid \mathbf W} $.
    \item [Lowbit SGD:] As stated in \eqref{eq:errorsgd} weight update in each iteration $k$ involves two kinds of errors, the error in computing gradient $\mathbf e_{1k}$, and the error in computing the update $\mathbf e_{2k}$. We wonder the gradient $\mathbf g_k$ needs to be redefined using a computationally more meaningful way such as  $\mathbf g_k(\eps) \approx {\mathcal L(\mathbf w_k) - \mathcal L(\mathbf w_k +\eps) \over \eps  } $ for a computationally meaningful $\eps$.
    \item[Fusion:] We wonder if the \emph{train large then compress} \citep{li2020train} can be regarded as an adaptive method to fuse these two steps. In other words, the number of model parameters need to be updated during training to combine training and compression into a single framework. Start with a large $\mathbf w_0$ and in each SGD update $\mathbf w_k$ not only updates the weight values but also updates $\mathrm{dim}(\mathbf w_k)$, for instance $\mathrm{dim}(\mathbf w_k) =  \one(\mathbf w_{k-1}>\epsilon)$.
   \item[Meta Size:] Dropout provides training many sparse models. On the other hand each data carries a weight about its true dimension through \eqref{eq:datadim} which is difficult to compute. Averaging over the sparse models from dropout instead of data can estimate the required dimension during training. Suppose each iteration consist of a dropout with $d_k$ activated neurons $d_k = \one(\mathbf w_k\neq 0), $ and $\hat d\approx {1\over K} \sum_k g(d_k).$ 
   \item[Meta SGD:]We propose to explore a meta SGD where the weights and model complexity are updated simultaneously until matching the model dimension with the data dimension to combine training and compression in a single framework. In each iteration i)  update weights $\mathbf w_k = g_1(\mathbf w_{k-1})$ ii) update model dimension $\mathrm{dim}(\mathbf w_k) = g_2(\mathbf w_{k-1}), $ iii) estimate data dimension $\hat d_k = g_3 (f_{\mathbf w_k}),$ take a gradient step to on the dimension space to bring them closer as the function of the other two dimensions $g_4(\hat d_k, \mathrm{dim}\{\mathbf w_k) \}$.    
   \item[Meta update:]  We wonder how to find  an optimal update strategy in \eqref{eq:gen_update} which is explicit enough to be implemented for a wide class of models, and at the same time general enough to be used for a large class of models. As a special case  $\mathbf w_{k+1} = h(\mathbf w_{1:k}, \mathbf  g_{1:k}, k, \mathcal F)$  can be refined to choosing the proper scheduling. Different depth of ResNets and different depth of Transformers are scheduled differently, so inherently  $\eta_k$ is  $\eta_k(f)$. 
   \item[Complexity:] Define the complexity measure $g(n, \mathcal F)$ and assumptions on the hypothesis class $\mathcal F$ beyond the classical theory to obtain tight bounds for in-domain and out-of-domain generalization bounds in \eqref{eq:vc_dim}, and \eqref{eq:rademacher} respectively that ideally satisfies $\Pr\{\epsilon_\Dc(f) \leq \epsilon_\Sc(f)+g(n, \mathcal F)\}\geq 1-\delta$ while satisfying $g(n, \mathcal F)\leq 1$ and $g(n, \mathcal F)\to 0$. 
   \item[Transfer:] We wonder if transferring learning from the source distribution $p_{\mathcal S} $ to the target $p_{\mathcal T}$ distribution needs to be re-formalized so that they cover pre-training (source) and fine-tuning (target) while conditions are  i) formally meaningful ii) practically verifiable iii) exhibit tight bounds for out-of-domain generalization in \eqref{eq:rademacher}.     
   \end{itemize}

\bibliography{main}
\bibliographystyle{CUP}
\end{document}